# RESEARCH ON SELF-CROSS-TRANSFORMER MODEL OF POINT CLOUD CHANGE DETECTION

Xiaoxu Ren[1,2], Haili Sun[1,,2], Zhenxin Zhang[1,2]

[1] Key Laboratory of 3D Information Acquisition and Application, MOE, Capital Normal University, Beijing 10048, China
[2] Colleage of Resource Environment and Tourism, Capital Normal University,Beijing 100048, Chnia

**ABSTRACT:**

With the vigorous development of the urban construction industry, engineering deformation or changes often occur during the construction process. To combat this phenomenon, it is necessary to detect changes in order to detect construction loopholes in time, ensure the integrity of the project and reduce labor costs. Or the inconvenience and injuriousness of the road. In the study of change detection in 3D point clouds, researchers have published various research methods on 3D point clouds. Directly based on but mostly based on traditional threshold distance methods (C2C, M3C2, M3C2-EP), and some are to convert 3D point clouds into DSM, which loses a lot of original information. Although deep learning is used in remote sensing methods, in terms of change detection of 3D point clouds, it is more converted into two-dimensional patches, and neural networks are rarely applied directly. We prefer that the network is given at the level of pixels or points. Variety. Therefore, in this article, our network builds a network for 3D point cloud change detection, and proposes a new module Cross transformer suitable for change detection. Simultaneously simulate tunneling data for change detection, and do test experiments with our network.

## 1. INTRODUCTION

With the vigorous development of the construction industry, the engineering problems in the detection of construction are particularly significant, and we need to detect changes in different time periods in engineering applications. While many existing studies use 2D images for change detection, 3D point clouds bring some complementary information about height, which seems to be useful in the context of detecting construction ground aspects, since the main modifications occur on the height axis. Furthermore, spectral variability of the same object over time, differences in viewing angle between 2D image acquisitions, perspective and distortion effects can complicate change retrieval based on 2D data (Qin et al., 2016).

Although change detection in 3D data in engineering has been addressed in several studies, so far, experimental comparisons are lacking. To the best of our knowledge, the only comparative analysis (Shirowzhan et al., 2019) is still at a qualitative level and excludes deep learning (DL) methods, which represent the state-of-the-art in remote sensing. In this paper, we first design a framework for point cloud change detection, and propose a new cross-attention mechanism suitable for change detection. We conducted comparative tests on real datasets as well as on real datasets plus simulated datasets (capable of introducing in-construction changes). We then compare representative approaches from the state of the art, ranging from classical distance-based approaches to recent deep learning developments, in the context of aerial lidar surveys (ALS) in urban areas.

## 2. RELATED WORKS

In this section, we briefly review general methods for change detection in 3D point clouds. Existing 3D PC-based methods for urban environment change detection and characterization are reviewed. Although there are many ways to convert PCs to DSMs, we will not focus on these studies as they are not directly related to the scope of our paper.

Unlike 2D images organized in a regular pixel grid (2D grid), 3D point clouds generated by lidar are disordered and irregular, which makes it difficult to extract information from these data, and the time stamps between Comparing is even more difficult. In fact, the location and distribution of points can also be very different in unaltered regions. Therefore, some methods convert a 3D point cloud into a 2D matrix that provides elevation information in each pixel. These 2D grids are called digital surface models (DSMs).

The idea of this method to detect changes between two 3D point clouds is to compute the DSM of the two point clouds and directly subtract them to retrieve the difference. It was first used for architectural change extraction in Murakami et al. (1999). It is still frequently used due to its simplicity and quality of results, and this method is also commonly used in the Earth observation community (Okyay et al., 2019); the difference in results can also be segmented using the Otsu threshold algorithm, which is calculated by minimizing each Variance between categories, thresholded from histogram of values (Otsu, 1979); since DSM contains artifacts (e.g. due to interpolation in hidden parts or difficulty retrieving precise architectural boundaries) (Gharibbafghi et al., 2019) ; There are several ways to apply more complex pipelines to derive more precise and finer-grained changes than positive or negative changes. For example, Choi et al. (2009) et al. use DSM difference to identify change regions, and then segment each change region by filtering and grouping; after empirical thresholding of DSMd results, it can also be based on the size of the remaining pixel clusters, Height and shape are used to select 3D building variations (Dini et al., 2012); from 3D point clouds, DSM and digital terrain models (DTM) can be extracted relying on ground points. Teo (Teo et al., 2012) et al. retrieve and classify each object at each date using DSM difference and DTM, then the segmented objects can be compared between the two time periods to determine changes; Pang et al. (Pang et al., 2014) still extract architectural change candidates with DSM change thresholds based on DSM changes; DSM differences with basic thresholds or further refinements are still widely used for 3D urban monitoring (Warth et al., 2019) and post-disaster buildings In the change detection literature for damage assessment (Wang et al., 2020).

With the rise of deep learning methods in Earth observation, change detection in 2D imagery has also benefited from this advance. For example, applying a convolutional neural network (CNN) to RGB images can assess building damage due to earthquakes (Kalantar et al., 2020); this study compared three different architectures, and the best results were Siamese architecture (Siamese networks), Siamese networks are used for change detection or similarity computation between two inputs, thus, they have been heavily used in remote sensing applications (Shi et al., 2020); even in optical and synthetic aperture radar They can also provide reliable results in the case of non-uniform inputs such as (SAR) images (Mou et al., 2017); since (Zhang et al., 2019) aims to use airborne (ALS) and photogrammetry to generate The dual-temporal multimodal 3D information of the point cloud, they chose the Siamese architecture so that it is fed into one branch DSM (either directly into the DSM difference or into two channels), and in the other branch into the corresponding RGB quadrature image, in their study they also calculated the region of change using only the DSM information.

Another 3D change detection method relies directly on raw point clouds. First, Girardeau-Montaut et al. (2005) proposed a cloud-to-cloud (C2C) comparison based on Hausdorff point-to-point distances and octree subdivisions of PCs to speed up computation; Lague et al. (2013) developed a A more refined method for measuring the average surface variation along the normal. Extract surface normals and orientations at a consistent scale based on local surface roughness. This approach is called Multiscale Model-to-Model Cloud Comparison (M3C2); the second technique allows distinguishing between positive and negative changes, which is not possible in C2C methods and, more importantly, requires less computational effort (Shirowzhan et al., 2019).

There are also semantic-based approaches. Among them, Awrangjeb et al. (2015) first extract boundaries from lidar data and aerial images, extract 2D footprints of buildings, and then compare the footprints to highlight changes on 2D maps; (Xu et al., 2015) also propose to segment each point cloud to extract buildings, and then create a 3D surface disparity map by computing the point-to-plane distance between points in the first set and the closest plane in the second set.

While rasterizing 3D data into a 2D elevation matrix (known as a digital surface model (DSM)) can be considered a valuable solution, this rasterization process results in some loss of information because only the The highest point, making it difficult to process such unstructured point cloud data directly with standard tools for 2D images. While existing studies that directly deal with 3D point clouds rely on hand-crafted features or distance computations, very few deep learning models are used to directly deal with the change detection representation problem of raw 3D point clouds. In recent years, deep learning methods have achieved good results in remote sensing and other fields. Therefore, designing a high-precision deep network model that can directly process 3D point clouds can provide a new method for point cloud change detection. This leads us to the next chapter.

## 3. METHOD

### 3.1 Background

To address the problem of change detection and representation in 2D images, recent studies propose to use deep Siamese fully convolutional networks (FCNs). It is contained in a common encoder-decoder network with skip connections. To extract features, both images will pass through the encoder part, which consists of two branches of the two images. Each branch is a series of traditional convolution and pooling layers to extract data information at several scales. The particularity of the Siamese network is that at each step of pooling, the difference of the extracted features of the two branches is maintained and cascaded at the corresponding scale in the decoder part (Daudt et al., 2018). If the data are very similar, the two branches of the encoder part can have shared weights to extract features in the same way. When the data are significantly different, for example, if images come from two types of sensors (such as optical and radar sensors), the weights may be independent, resulting in a so-called pseudo-Siamese network (Zhan et al., 2017).

To address the 3D part of the problem, we propose to rely on deep networks capable of performing semantic segmentation directly on the PC. To this end, we consider the core module of the recent Point transformer (Zhao et al., 2021), a network that achieves very good results on segmentation and classification tasks. In a neural network similar to a 2D image encoder, the principle is to apply an attention mechanism, adapting this operation to a 3D point cloud. The author of Point transformer implemented different types of networks inspired by traditional networks in natural language processing and 2D image transformers, applied them to point clouds, and added the position codes that point clouds themselves have. At the same time, we have made further improvements to the Point transformer, and proposed a new module Cross-Transformer. This cross-attention mechanism realizes attention calculations across point clouds and is more suitable for change detection tasks.

### 3.2 Our framework

To extend the Siamese principle to 3D point clouds, we here propose to embed a modified Point transformer architecture into a deep Siamese network, where point clouds from two different time periods will pass the same encoding with shared weights device. Similar to common encoder-decoders with skip connections, at each scale in the decoding part we concatenate the differences in extracted features associated with the corresponding encoding scale (see Figure 1). In practice, the computation of such feature differences is not noticeable, since point clouds do not contain the same number of points and are not defined at the same locations, even in unchanged regions. We perform feature transfer decoding by retrieving the feature of the corresponding spatial point in the first point cloud against this difference in the neighboring points of the point in the second point cloud, and adding it to the original point cloud.

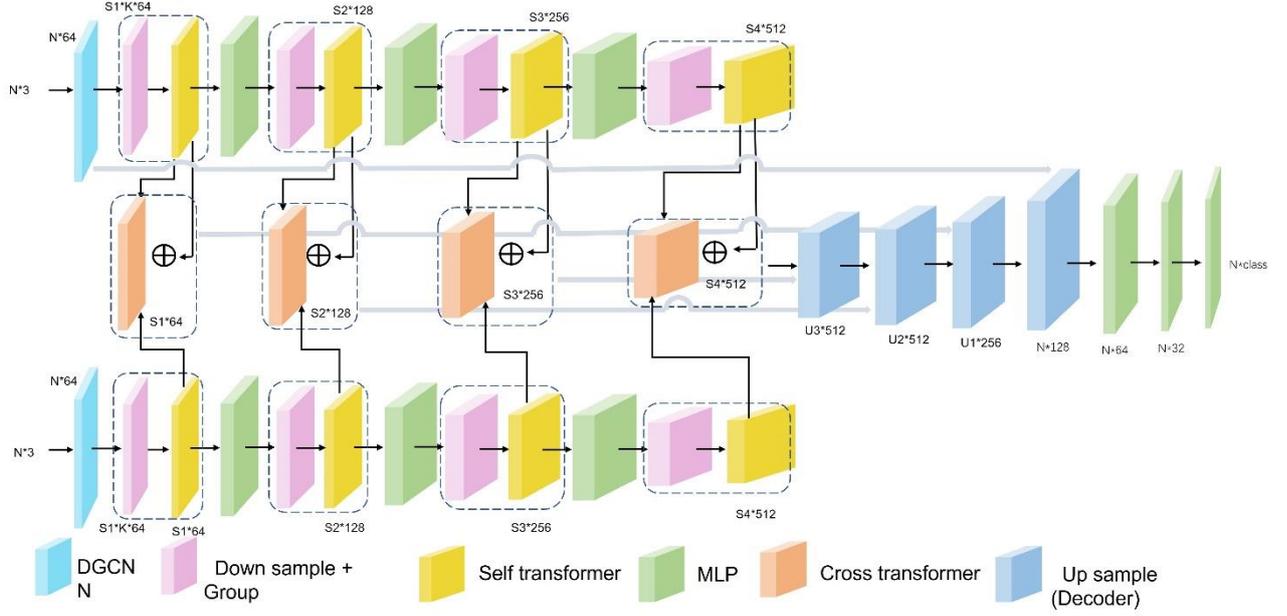

Figure 1. Our network framework diagram

In this method, we combine the Self transformer and the Cross transformer in the Point transformer, use the advantages of the Self transformer module and the Cross transformer module to extract the descriptors with resolution of the local point cloud, and use the dynamic graph convolution as the feature embedding of the transformer , and finally achieve point-by-point level change detection of point clouds.

In this section, we refer to (Wang et al., 2019) to operate dynamic graph convolutions as feature embeddings of the Self-transformer module. Different from (Zhao et al., 2021) directly using MLP as Transformer, through experiments, we verified that using dynamic graph convolution as its feature embedding has better results. Because it dynamically updates the graph structure, it can learn to semantically group points by dynamically updating the relationship graph from one layer to another, not only limited to local features with similar distances.

In this structure, the dynamic calculation graph structure is the essence of this module. It is beneficial to recompute the graph using the k nearest neighbor points in the feature space produced by each layer. This is a key difference between this method and graph CNNs that work on fixed input graphs. With dynamic graphic updates, the receptive field range gradually increases to be as large as the point cloud diameter. At each layer, construct a different graph structure $G^l = (V^l, E^l)$, where the edge structure of the lth layer is $(i, j_{i1}), ..., (i, j_{ikl})$, making $x_{j_{i1}}^{(l)}, ... x_{j_{ikl}}^{(l)}$ are the points closest to $x_i^{(l)}$, in other words, the architecture learns how to build the graph used in each layer G, rather than having it as a fixed constant constructed before evaluating the network. In the process of implementation, we calculate a paired distance matrix in the feature space, and then take the closest k points for each point, instead of calculating the surrounding adjacent points according to the fixed distance according to the coordinates, which is more conducive to the proximity of the distance into semantic proximity.

In this section, we design a self-transformer-based feature encoding module to process point clouds and effectively extract 3D features of scene point clouds. The Self transformer encoder-decoder module consists of an encoder and a decoder (Figure 1). We adopt a twin structure in the encoder, and the two branches share the encoder with four layers of SA layers and Self transformer layers. Our network is upsampled and downsampled using Pointnet++ (Qi et al., 2017). At the same time, position coding is added.

This paper uses a vector attention operator different from scalar attention, which can be verified in (Zhao et al., 2021). In the paper (Zhao et al., 2021), the authors used scalar attention. Compared with scalar attention, the calculation of attention weight in vector attention is different. Vector attention can adjust the attention between each feature channel.We use a subtractive

relationship and add the positional encoding σ to the attention vector and the transformed features:

$$y_i = \sum_{x_j \in X(i)} \rho\left(\beta(\varphi(x_i) - \omega(x_j) + \sigma)\right) \odot (\alpha(x_j) + \sigma) \quad (1)$$

$$F_{out} = F_{in} + y_i \quad (2)$$

Here the relation function uses subtraction, and $\beta$ is a mapping function which is an MLP with two linear layers and a ReLU non-linearity, generating attention vectors for feature aggregation. where $y_i$ is the output attention feature, $F$ is the total output feature, and $\varphi$、$\omega$, and $\alpha$ are point-wise feature transformations, such as linear projection or MLP. $\sigma$ is a position encoding function and $\rho$ is a normalization function such as softmax. Here the subset $X(i) \subseteq X$ is a set of points in the local neighborhood (specifically the k nearest neighbors) of $x_i$. The difference between us and the Point transformer is that we use Dynamic graph cnn (DGCNN) (Wang et al., 2019) as the feature embedding, and add the L1 norm to ρ for normalization.

Inspired by the self-attention mechanism in the previous section, the biggest innovation in this paper is to propose a cross-attention mechanism for change detection, which we call Cross-Transformer. We still use Equation 1 in this section.

Similar to natural language processing tasks, in the transformation, we use the linear transformation features of $x_i$ as the query matrix Q , the linear transformation features of k points in the local neighborhood of $x_i$ as the key matrix, and the k points in the local neighborhood of $x_i$ are different The matrix of values generated by the linear transformation as V.

The difference from self-attention is that when we calculate attention, we use the fusion between two point clouds. Here we still use subtraction to calculate the attention score. This has been verified in our experiments. Using subtraction is more effective than other operations, because we are doing a change detection task, and the difference between the two is more prominent. It is reflected in the calculation that Q and K come from different point clouds, and KNN is used to find adjacent K points, and the information is fused to find the difference to calculate the attention score of the corresponding position between the two point clouds, and finally I Attention score weights are assigned to K surrounding point features.

## 4. EXPERIMENTS

### 4.1 Dataset

We use two sets of data for training and testing in this paper. One data set is an open-pit iron mine located on September 12, 2017 and January 12, 2017 in Liaoning Province, China. It has an area of 0.6 square kilometers and a maximum depth of 170 meters. Another data set is from the shield tunnel in Huaide County, Jilin Province, China, from 2021 and 2022 respectively. In this set of data, because the amount of change is far from enough for the tunnel to be used for training, So we added our own simulations to this set of data in addition to some of its own changes. These include the shedding of wires, the construction of internal catenary, and the construction of infrastructure to simulate changes in construction. The data simulation is real and can be used as our research data. In the test data set, we used 3/1 of the two data sets for testing.

### 4.2 Results

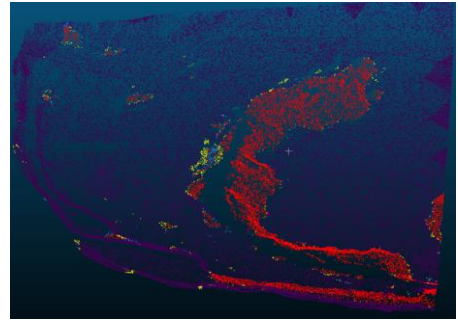

Figure 2. mine test results

As shown in Figure 2 above, the ground subsidence change detection is performed in Dataset 1. The red part in the figure represents the correct detection in the change category, the purple represents the correct detection in the unchanged category, the blue part is missing in the change type, and the yellow part It is omitted in the invariant type, and it can be seen from the results that our result shows a good detection effect. From our metrics in Table 1 we can see that the accuracy metrics are better.

| mine | mean evaluation | | | | | |
|---|---|---|---|---|---|---|
|  | OA | mrecall | mprecision | mf1score | mIoU | % |
| testdata | 98.05 | 97.44 | 95.76 | 96.34 | 93.46 | |

Table 1. Change detection results of mine data set

| tunnel | mean evaluation | | | | |
|---|---|---|---|---|---|
| | OA | mrecall | mprecision | mf1score | mIoU  % |
| test1 | 99.51 | 96.96 | 97.19 | 97.08 | 94.43 |
| test2 | 99.43 | 99.05 | 94.59 | 96.70 | 93.77 |

Table 2. Change detection results of tunnel dataset

In our tunnel change detection, we divide the test tunnel into two data for easy observation. In Figure 3 in the figure below, in Table 2 we counted the detection results, and showed good results in both data sets 1 and 2. In Figure 3, we compared with the real value, although the internal facilities of the tunnel are complex , but most of them can still be detected, such as the slight changes in the increase of wire changes under the water pipe on the right side of the tunnel wall in test1 can still be detected.

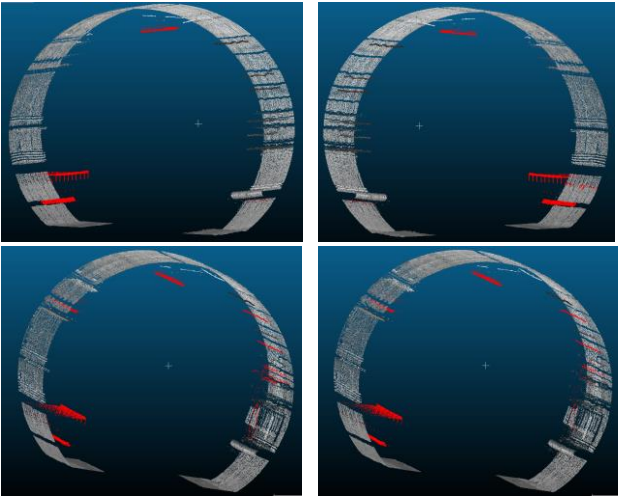

Figure 3. the two pictures on the left are the true values of the two data sets, and the right is the model detection results (red is the change, the two pictures on the top are test1, and the two pictures on the bottom are test2)

## 5. DISCUSSION

We tested each of the three test data sets. Although we still have a comparison between our method and many traditional methods, we have a comparison of our model simulated in other (de Gélis et al., 2023) papers. Compared with the city datasets, our model is much more optimized than traditional algorithms. In our model, we observe that most of the change detection can be detected, such as in the ground subsidence of the mining field, we study the change in elevation, which is similar to the urban change detection (de Gélis et al., 2023), and in more complex tunnels, our network can still detect most of the changes, but for some changes, because the structure is too complex and similar, the changes that are relatively close to the tunnel wall have not been detected. It is worthy of our attention, and it is also the room for further improvement of the network.

## 6. CONCLUSION

In this article, we propose a new point cloud change detection method, using a deep neural network, combined with a dynamic graph convolutional network and a Transformer model, to build a change detection network and extract descriptors with strong discriminative power. At the same time, based on the change detection task, this paper designs a new network module Cross transformer, and proposes a cross-attention mechanism, which can calculate the attention score between point clouds, improve the local fusion performance of the two point clouds, and more accurately find change area.

We apply deep learning to the field of 3D point cloud change detection, but the label production of the data set still relies on manual work, and the results are no longer presented in the form of patches, and point-level change detection has been realized. Our future work will be more Focus on the model optimization innovation of 3D point cloud, and hope that more scholars will do and provide more data sets on the task of 3D point cloud change detection.

## REFERENCES

Awrangjeb, M., Fraser, C. S., Lu, G., 2015. Building change detection from LiDAR point cloud data based on connected component analysis. *ISPRS annals of the photogrammetry, remote sensing and spatial information sciences, 2, 393-400.*

Choi, K., Lee, I., Kim, S., 2009. A feature based approach to automatic change detection from LiDAR data in urban areas. *Int. Arch. Photogramm. Remote Sens. Spat. Inf. Sci*, *18*, 259-264.

Daudt, R. C., Le Saux, B., Boulch, A., 2018, October. Fully convolutional siamese networks for change detection. In *2018 25th IEEE International Conference on Image Processing (ICIP)* (pp. 4063-4067). IEEE.

de Gélis, I., Lefèvre, S., Corpetti, T., 2023. Siamese KPConv: 3D multiple change detection from raw point clouds using deep learning. ISPRS Journal of Photogrammetry and Remote Sensing, 197, 274-291.

Dini, G. R., Jacobsen, K., Rottensteiner, F., Al Rajhi, M., Heipke, C., 2012. 3D building change detection using high resolution stereo images and a GIS database. The International Archives of the Photogrammetry, Remote Sensing and Spatial Information Sciences; XXXIX-B7, 39, 299-304.

Gharibbafghi, Z., Tian, J., Reinartz, P., 2019. Superpixel-Based 3D Building Model Refinement and Change Detection, Using VHR Stereo Satellite Imagery. In Proceedings of the IEEE/CVF Conference on Computer Vision and Pattern Recognition Workshops (pp. 0-0).

Girardeau-Montaut, D., Roux, M., Marc, R., Thibault, G., 2005. Change detection on points cloud data acquired with a ground laser scanner. International Archives of Photogrammetry, Remote Sensing and Spatial Information Sciences, 36(3), W19.

Kalantar, B., Ueda, N., Al-Najjar, H. A., Halin, A. A., 2020. Assessment of convolutional neural network architectures for earthquake-induced building damage detection based on pre-and post-event orthophoto images. Remote Sensing, 12(21), 3529.

Lague, D., Brodu, N., Leroux, J., 2013. Accurate 3D comparison of complex topography with terrestrial laser scanner: Application to the Rangitikei canyon (NZ). ISPRS journal of photogrammetry and remote sensing, 82, 10-26.

Mou, L., Schmitt, M., Wang, Y., Zhu, X. X., 2017, March. A CNN for the identification of corresponding patches in SAR and optical imagery of urban scenes. In 2017 Joint Urban Remote Sensing Event (JURSE) (pp. 1-4). IEEE.

Murakami, H., Nakagawa, K., Hasegawa, H., Shibata, T., Iwanami, E., 1999. Change detection of buildings using an airborne laser scanner. ISPRS Journal of Photogrammetry and Remote Sensing, 54(2-3), 148-152.

Okyay, U., Telling, J., Glennie, C. L., Dietrich, W. E., 2019. Airborne lidar change detection: An overview of Earth sciences applications. Earth-Science Reviews, 198, 102929.

Otsu, N., 1979. A threshold selection method from gray-level histograms. IEEE transactions on systems, man, and cybernetics, 9(1), 62-66.

Pang, S., Hu, X., Wang, Z., Lu, Y., 2014. Object-based analysis of airborne LiDAR data for building change detection. Remote Sensing, 6(11), 10733-10749.

Qi, C. R., Yi, L., Su, H., Guibas, L. J., 2017. Pointnet++: Deep hierarchical feature learning on point sets in a metric space. Advances in neural information processing systems, 30.

Qin, R., Tian, J., Reinartz, P., 2016. 3D change detection–approaches and applications. ISPRS Journal of Photogrammetry and Remote Sensing, 122, 41-56.

Shi, W., Zhang, M., Zhang, R., Chen, S., Zhan, Z., 2020. Change detection based on artificial intelligence: State-of-the-art and challenges. Remote Sensing, 12(10), 1688.

Shirowzhan, S., Sepasgozar, S. M., Li, H., Trinder, J., Tang, P., 2019. Comparative analysis of machine learning and point-based algorithms for detecting 3D changes in buildings over time using bi-temporal lidar data. Automation in Construction, 105, 102841.

Teo, T. A., Shih, T. Y., 2013. Lidar-based change detection and change-type determination in urban areas. International journal of remote sensing, 34(3), 968-981.

Wang, Y., Sun, Y., Liu, Z., Sarma, S. E., Bronstein, M. M., Solomon, J. M., 2019. Dynamic graph cnn for learning on point clouds. ACM Transactions on Graphics (tog), 38(5), 1-12.

Wang, X., Li, P., 2020. Extraction of urban building damage using spectral, height and corner information from VHR satellite images and airborne LiDAR data. ISPRS Journal of Photogrammetry and Remote Sensing, 159, 322-336.

Warth, G., Braun, A., Bödinger, C., Hochschild, V., Bachofer, F., 2019. DSM-based identification of changes in highly dynamic urban agglomerations. European Journal of Remote Sensing, 52(1), 322-334.

Xu, S., Vosselman, G., Oude Elberink, S., 2015. Detection and classification of changes in buildings from airborne laser scanning data. Remote sensing, 7(12), 17051-17076.

Zhan, Y., Fu, K., Yan, M., Sun, X., Wang, H., Qiu, X., 2017. Change detection based on deep siamese convolutional network for optical aerial images. IEEE Geoscience and Remote Sensing Letters, 14(10), 1845-1849.

Zhang, Z., Vosselman, G., Gerke, M., Persello, C., Tuia, D., Yang, M. Y., 2019. Detecting building changes between airborne laser scanning and photogrammetric data. Remote sensing, 11(20), 2417.

Zhao, H., Jiang, L., Jia, J., Torr, P. H., Koltun, V., 2021. Point transformer. In Proceedings of the IEEE/CVF international conference on computer vision (pp. 16259-16268).